\documentclass{midl} 

\title[Group-Attention Single-Shot Detector]{Group-Attention Single-Shot Detector (GA-SSD): Finding Pulmonary Nodules in Large-Scale CT Images}
\midlauthor{\Name{Jiechao Ma\nametag{$^{1,3}$}} \Email{majch7@mail2.sysu.edu.cn}\\
\Name{Xiang Li\nametag{$^{1}$}} \Email{lixiang651@gmail.com}\\
\Name{Hongwei Li\nametag{$^{2}$}} \Email{hongwei.li@tum.de}\\
\Name{Bjoern H Menze\nametag{$^{2}$}} \Email{bjoern.menze@tum.de}\\
\Name{Sen Liang\nametag{$^{3}$}} \Email{lsen@infervision.com}\\
\Name{Rongguo Zhang\nametag{$^{3}$}} \Email{zrongguo@infervision.com}\\
\Name{Wei-Shi Zheng\nametag{$^{1}$}} \Email{wszheng@ieee.org}\\
\addr $^{1}$ School of Data and Computer Science, Sun Yat-sen University, China. \\
\addr $^{2}$ Department of Computer Science, Technical University of Munich, Germany.\\
\addr $^{3}$ Infervision Inc, China. \\
}

\begin{document}

\maketitle

\begin{abstract}
Early diagnosis of pulmonary nodules (PNs) can improve the survival rate of patients and yet is a challenging task for radiologists due to the image noise and artifacts in computed tomography (CT) images. In this paper, we propose a novel and effective abnormality detector implementing the attention mechanism and group convolution on 3D single-shot detector (SSD) called group-attention SSD (GA-SSD). We find that group convolution is effective in extracting rich context information between continuous slices, and attention network can learn the target features automatically. We collected a large-scale dataset that contained 4146 CT scans with annotations of varying types and sizes of PNs (even PNs smaller than 3mm). To the best of our knowledge, this dataset is the largest cohort with relatively complete annotations for PNs detection. Extensive experimental results show that the proposed group-attention SSD outperforms the conventional SSD framework as well as the state-of-the-art 3DCNN, especially on some challenging lesion types.
\end{abstract}


\section{Introduction}
Lung cancer continues to have the highest incidence and mortality rate worldwide among all forms of cancers \cite{Bray}. Because of its aggressive and heterogeneous nature, diagnosis and intervention at the early stage, where cancer manifests as pulmonary nodules, are vital to survival \cite{siegel2018cancer}. Although the use of a new generation of CT scanners improves the detection of pulmonary nodules, certain nodules (such as ground-glass nodules, GGGN) are still misdiagnosed due to noise and artifacts in CT imaging. \cite{manning2004detection,hossain2018missed}. The design of a reliable detection system is increasingly needed in clinical practice.

Deep learning techniques using convolution neural networks (CNN) is a promising and effective approach to assisting lung nodule management. For example, Setio \cite{Setio2016Pulmonary} proposed a system for pulmonary nodule detection based on multi-view CNN, where the network is fed with nodule candidates rather than whole CT scans. Wang \cite{inbook} presented a 3D CNN model trained with feature pyramid networks (FPN) \cite{Lin2016Feature} and achieved the state-of-the-art on  LUNA16\footnote{https://luna16.grand-challenge.org/}. However, all these algorithms neither make use of the spatial attention across the neighboring slices nor introduce the attention mechanism for region of interest, because the regional distribution of target PNs and the non-PNs is highly unbalanced. Therefore, learning to automatically weight the importance of slices and pixels is essential in pulmonary nodules detection.

In this work, to address the problem of indiscriminate weighting of pixels and slices, we propose a lung nodule detection model called group-attention SSD (\emph{GA-SSD}), which leverages one-stage single-shot detector (SSD) framework \cite{liu2016ssd,fu2017dssd,luo20173d} and attention module with group convolutions. Firstly, a group convolution is added at the beginning of the GA module to weight the importance of input slices. Secondly, the attention mechanism is integrated into the grouped features to enhance the weight of nodule's pixels on a 2D image.

We evaluate the proposed system on our challenging large-scale dataset containing 4,146 patients. Different from existing datasets, the cohort contains eight categories of PNs including ground-glass nodules (GGNs) which are hard-to-detect lesions of clinical significance yet not usually included in conventional datasets.

\section{Related Works}
\textbf{Object Detection.} Recent object detection models can be grouped into one of two types \cite{liu2018deep}, two-stage approaches \cite{girshick2014rich,girshick2015fast,ren2015faster} and one-stage methods \cite{redmon2016you,liu2016ssd}. The former generates a series of candidate boxes as proposals by the algorithm and then classifies the proposals by convolution neural network. The latter directly transforms the problem of target border location into a regression problem without generating candidate boxes. It is precise because of the difference between the two methods, the former is superior in detection accuracy and location accuracy, and the latter is superior in algorithm speed. 

\noindent\textbf{Attention Modules.} The inspiration of attention mechanism comes from the mechanism of human visual attention. Human vision is guided by attention which gives higher weights on objects than background.  Recently, attention mechanism has been successfully applied in natural language processing \cite{vaswani2017attention,cho2014learning,sutskever2014sequence,yang2016hierarchical,yin2015abcnn} as well as computer vision \cite{fu2017look,zheng2017learning,sun2018multi}. Most of the conventional methods which solve the object detection problems neglect the correlation between proposed regions. The Non-local Network \cite{wang2018non} and the Relation networks \cite{hu2018relation} were translational variants of the attention mechanism and utilize the interrelationships between objects. In medical image analysis community, oktay \cite{oktay2018attention} introduced attention mechanism to solve the pancreas segmentation problem. Our method is motivated by these works, aiming at medical images, to find the inter-correlation between CT slices and between lung nodule pixels. 

\noindent\textbf{Group Convolution.} Group convolution first appeared in AlexNet\cite{krizhevsky2012imagenet}. To solve the problem of insufficient memory, AlexNet proposed that the group convolution approach could increase the diagonal correlation between filters and reduce the training parameters. Recently, many successful applications have proved the effectiveness of group convolution modules such as channel-wise convolution including the Xception \cite{43022,szegedy2016rethinking} (Extreme Inception) and the ResNeXt\cite{xie2017aggregated}.

\section{Methodology}

In this section, we present an effective 3D SSD framework for lung nodule detection because the detection of lung nodules relies on 3D information. The proposed framework has two highlights: the attention module and grouped convolution. We call this new model group attention SSD (GA-SSD) because we integrate group convolution with attention modules.

\subsection{Overall Framework}
The proposed 3D SSD shares the basic architecture of the classic SSD \cite{liu2016ssd}. And the network structure of the \emph{GA-SSD} can be divided into two parts: the medical image loading sub-network for reading CT scans (pre-process) and the backbone sub-network for feature extraction. Specifically, we use the deeply supervised ResNeXt \cite{he2016deep} structure as the backbone, and add \emph{GA} structure to both pre-process sub-network and backbone sub-network respectively.

\begin{figure}
	\centering
	\includegraphics[width=15cm,height=7cm]{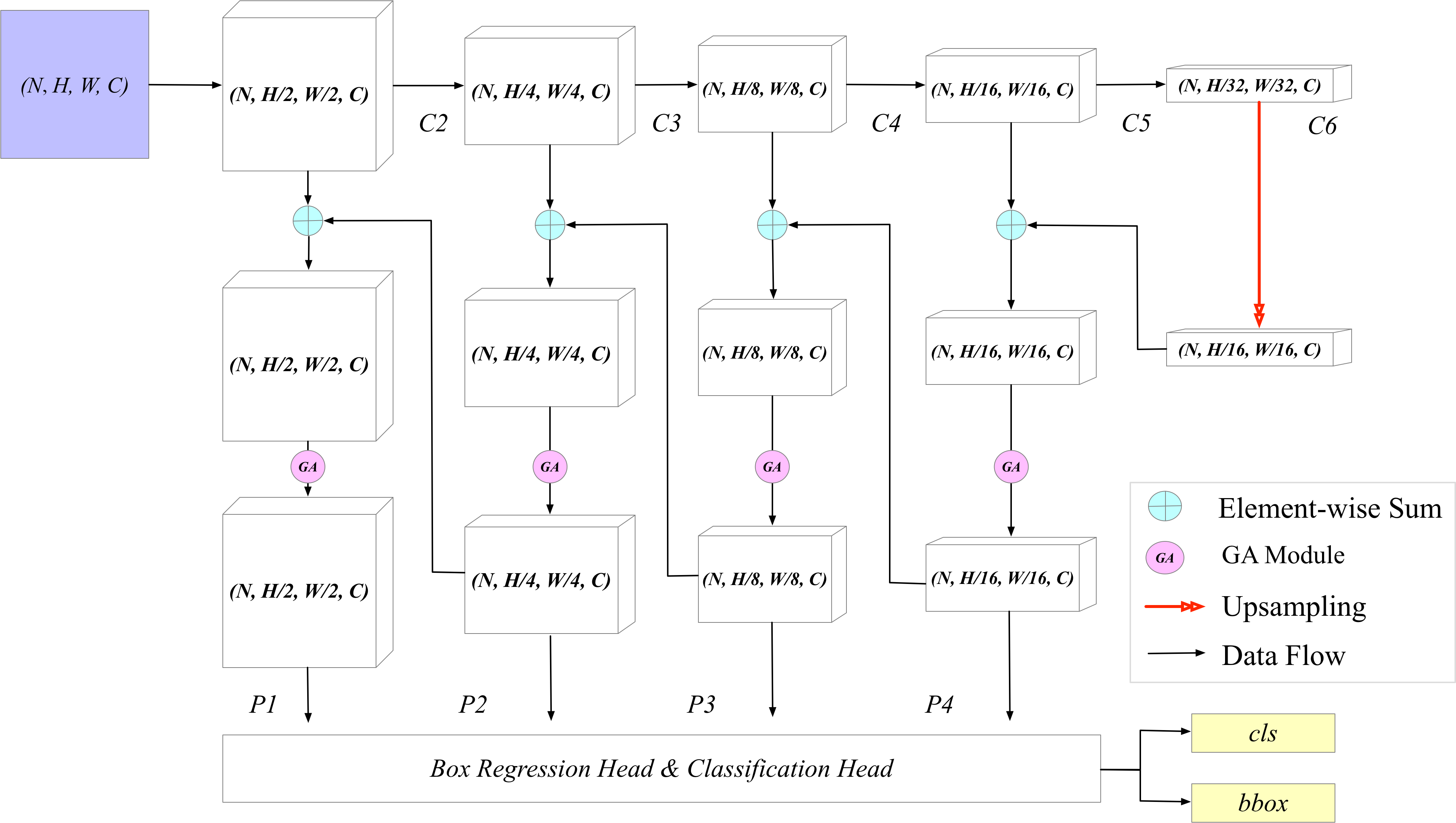}
	\caption{The architecture of SSD framework with FPN and attach the GA module.}
	\label{fig:GA_module1}
\end{figure}
\subsection{SSD Architectures}
The classic SSD network was established with VGG16 \cite{simonyan2014very}. Compared with Faster RCNN \cite{ren2015faster}, the SSD algorithm does not generate proposals, which greatly improves the detection speed (to handle the large-scale dataset). The basic idea of SSD is to transform the image into different sizes (image pyramid), detect them separately, and finally synthesize the results.

In this work, the original ResNeXt structure is modified into FPN-like layers (Figure \ref{fig:GA_module1}). In order to detect the small object, four convolution layers (P1, P2, P3, P4) are added to construct the different size candidate boxes. And these outputs (boxes) are optimized by two different losses (regression, classification), resulting in one class confidence (each default box generates several class confidences) and one output regression localization (each default box generates four coordinate values $(x,y,w,h)$). In order to utilize the three-dimensional information between lung CT slices, we modify the basic structure of SSD network to improve the performance of the detection framework. The backbone network uses 3D convolution with the same padding instead of the conventional 2D convolution layer. And the Rectified linear unit (ReLU) is employed as activation functions to the nodes. Additionally, we apply dropout regularization\cite{srivastava2014dropout} to prevent complex node connections.

\subsection{\emph{GA} Modules}
\begin{figure}
	\centering
	\includegraphics[width=15cm,height=6cm]{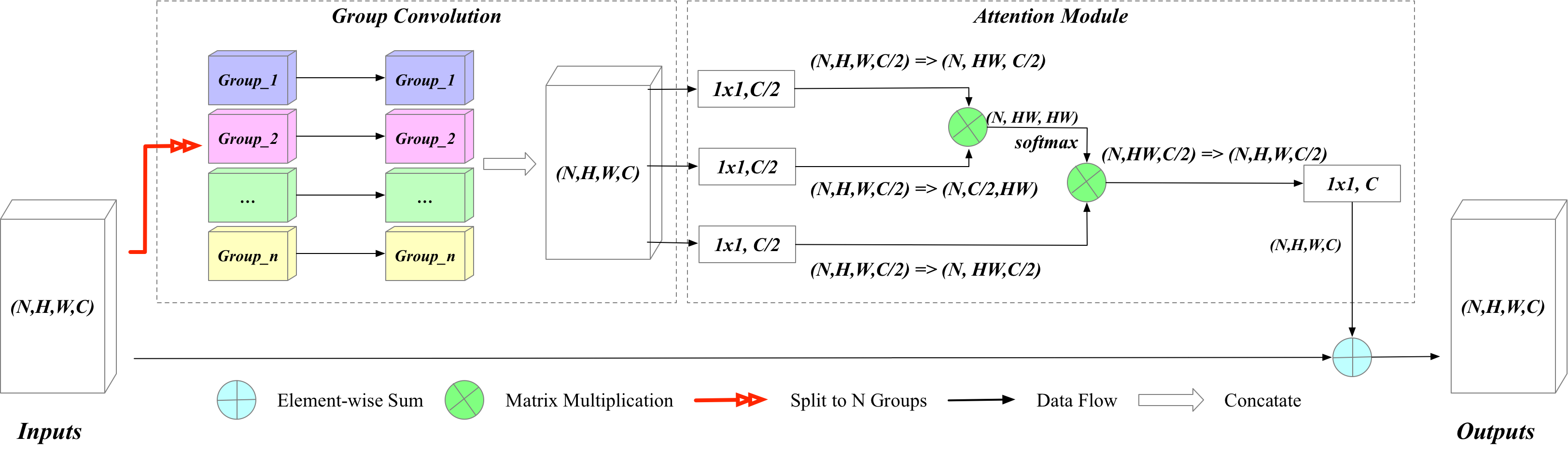}
	\caption{The architecture of GA module, it can be divided into two parts: The former enforces a sparsity connection by partitioning the inputs (and outputs) into disjoint groups. The latter use the concatenated groups to find the non-local information.}
	\label{fig:GA_module}
\end{figure}

To imitate the usual viewing habits of radiologists who usually screen for nodule lesions in 2-D axial plane at the thin-section slice, we propose a new medical imaging group convolution and attention-based network (GA module (Figure \ref{fig:GA_module})) to tell the model which slices of the patient to focus on and automatically learn the weight of these slices. See figure \ref{fig:GA_module}, assume that the input feature map is $(N, H, W, C)$, which means the channel is $C$, the batch size is $N$, the width and height are $H, W$, respectively. Suppose the number of group convolutions is $M$ (we use $M=9$ by default). So the operation of this group convolution is to divide channels into $M$ parts. Each group corresponds to $C/M$ channels and is convoluted independently. After each group convolution is completed, the output of each group is concatenated as the output channel $(N, H, W, C)$. And the attention mechanism based on sequence generation can be applied to help convolutional neural networks to focus on non-local information of images to generate weighted inputs of the same sizes as the original inputs.

The GA behavior in Eq.(1) is due to the fact that all pixels are considered in the operation. $f(x_i, x_j)$ is used to calculate the pairwise relationship between target $i$ and all other associated pixel $j$. This relationship is as follows: the farther the pixel distance between $i$ and $j$ is, the smaller the $f$ value is, indicating that the $j$ pixel has less impact on $i$. $g(x)$ is used to calculate the eigenvalues of the input signal at the $j$ pixel. $C(x)$ is a normalized parameter. In figure \ref{fig:GA_module}, we use three $1\times 1$ convolution layer to get corresponding features. Then use the softmax function ($f(x)$) and gaussian function $(g(x))$ to get the attention information.

$$y_i=\frac{1}{C(x)}\sum\limits_{\forall j}f(x_i,x_j)g(x_j).\eqno(1)$$

Our approach improves parameter efficiency and adopts group convolution and attention module, where group convolution acts to find the best feature maps (i.e., highlight several slices from the input CT scans), and the attention module acts to find the location of the nodule (i.e., the size and shape of the target nodule in a specific feature map).

In addition, due to the simplicity and applicability of GA module, it can be easily integrated into a standard CNN architecture. For example, we apply this module not only to the data loading sub-network but also the feature extraction stage of the network, which allows the model to automatically and implicitly learn some correlated regions of different features, and focuses on areas that the model needs to focus on.

\section{Experiments and Results}

\subsection{A Large-scale Computed Tomography Dataset}
\begin{figure}[!htb]
	\centering
	\includegraphics[width=12cm,height=4cm]{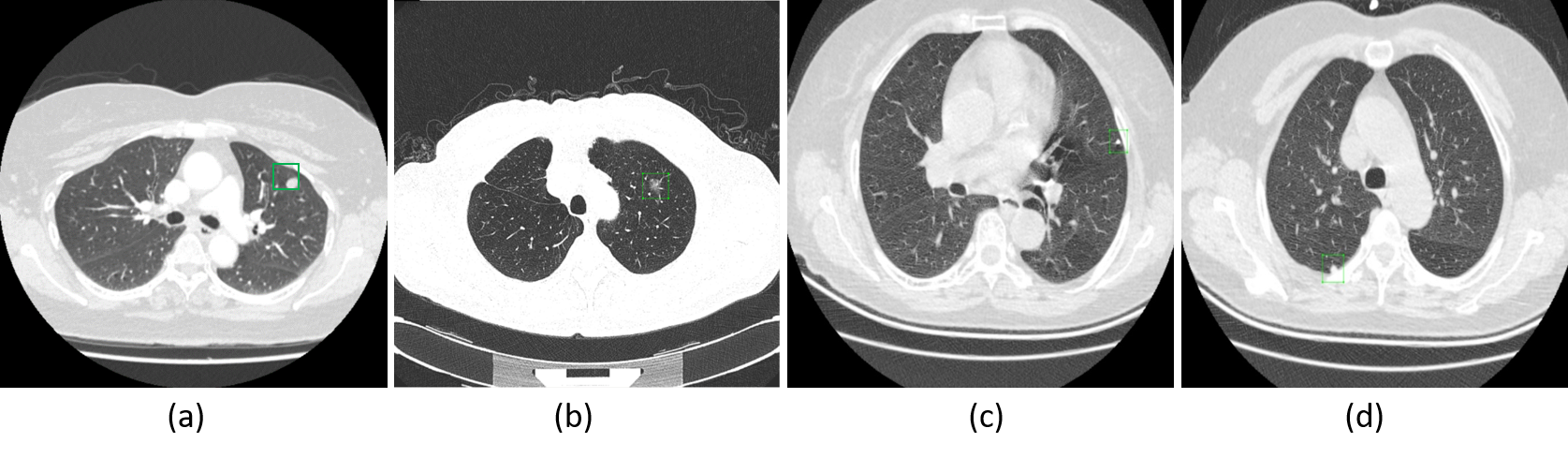}
	\caption{Sample CT images of the dataset used in evaluation of our deep learning model.(a) solid nodules, (b) subsolid nodules, (c) calcified nodules, and (d) pleural nodules}
	\label{fig:nodules}
\end{figure}

A cohort of 4146 chest helical CT scans was collected from various scanners from several centers in China. Each chest CT scan contains a sequence of slices. Pulmonary nodules were labeled by experienced radiologists after evaluating their appearance and sizes. They are divided into eight categories: \emph{calcified nodule with two different sizes, pleural nodule with two different sizes, solid nodules two different sizes, ground-glass nodule divided into pure-GGN and sub-solid nodules (mixed-GGN)} as shown in (Figure \ref{fig:nodules}). To the best of our knowledge, the current dataset is the largest cohort for PNs detection and with eight categories and varied sizes of annotated PNs. The detection of ground-glass nodules is important and challenging in clinical practice; however, it is not included as a part of the PNs detection task in conventional datasets such as LUNA16.

The images in our study were acquired by different CT scanners with Philips, GE, Siemens, Toshiba. All the chest CT images were acquired with the axial plane at the thin-section slice spacing (range from 0.8 to 2.5 mm). The regular dose CT scans were obtained at 100 kVp-140 kVp, tube current greater than 60mAs, $512*512$ pixel resolution; and the low-dose CT images were obtained that tube current less than 60mAs with all other acquisition parameters the same as those used to obtain the regular dose CT. For our experiment, we randomly selected $80\%$ of patients as training set and $20\%$ of patients as testing set. Gradient updates were computed using batch sizes of 28 samples per GPU. All models were trained using the SGD optimizer, implemented batch normalization, and data augmentation techniques. Lung nodule detection performance was measured by CPM\footnote{The code is opensource: https://www.dropbox.com/s/wue67fg9bk5xdxt/evaluationScript.zip}. The evaluation is performed by measuring the detection sensitivity of the algorithm and the corresponding false positive rate per scan. This performance metric was introduced in \cite{setio2017validation}.


\subsection{Effectiveness of \emph{GA} Module for Data Load}
Table~1 investigates the effects of different input methods. The baseline (multi-channel) used continuous slices as input and our approach added the GA module to this input mode. Using the GA module, our approach improved the CPM from 0.615 to 0.623. When we changed the 2.5D input  (multi-channel 2D) to 3D volume input by reshaping the dimensions, the results improved by (+0.018) vs (+0.008). These results verified the effectiveness of the GA module for data pre-processing. In addition, our approach works better on 3D data than on 2D data. 
\begin{table}[htbp]
\floatconts
  {tab:example}%
  {
\caption{Comparision of the input method with GA module.}
\label{table:mytab}
}
  {\begin{tabular}{l | c | c   c  c c c c c c}
  \hline
  \bfseries Input Method & \bfseries 2.5-D &  \bfseries 3-D  \\
  \hline\hline
  Multi-channel   &  0.615 &  0.654\\
  GA module   & \bfseries 0.623  & \bfseries 0.672\\
  \hline
  \end{tabular}}
\end{table}

\subsection{Effectiveness of \emph{GA} Module for Capturing Multi-scale Information}
For detecting small objects, feature pyramid is a basic and effective component in the system. In general, small lung nodules are challenging for detectors, this is because there are numerous small nodules which are around $20*20$ pixels in the $512*512$ image data, making it difficult to localize. Thus, FPN is an important component in our framework for detecting small nodules.

In order to better investigate the impact of FPN on the detector, we conducted comparison experiments on a fixed set of feature layers. From Figure \ref{fig:GA_module1}, we can see that on the basis of the original SSD (C2, C3, C4, C5, C6), the feature map of the latter layer uses upsampling to enlarge the size, and then adds with the former layer (original FPN). In the GA-FPN version, we use the GA module of the feature map as shown in Figure \ref{fig:GA_module} to get the weight between feature maps. We choose to calculate the candidate box (like SSD) for several of the layers on the FPN. The lower layer, such as the P1 layer, has better texture information, so it is associated with good performance on small targets that can be identified by the detector. The higher level has stronger semantic information, and is associated with better results for the category classification of nodules. For the sake of simplicity, we do not share feature levels between layers unless specified.

Table~2 compares the effect of FPN and GA-FPN across the feature maps. According to Table~2 (left) results, using a three-layer (P4, P3, P2) model as a baseline, when a feature is used at a very early stage (like P1), it brought more false positives and harmed the performance (drops from 0.654 to 0.473). However, compared to the output of the P4 layer alone (0.680), the model with relatively lower information of the P3 layer gained better performance.

According to Table~2 (right) results, the framework improved the overall performance across feature layers. That the framework performance improved from 0.696 (best in FPN ) to 0.721 (best in GA-FPN) validated our conjecture that using the GA module could help the model learn more important feature layers.


\begin{table}[htbp]
\floatconts
  {tab:example}%
  {\caption{Comparision of the feature extraction with GA module.}}%
  {\begin{tabular}{c | c  c   c  c |c  c c c c c}
  \hline
  \bfseries Feature & \multicolumn{4}{|c|}{\bfseries FPN}  & \multicolumn{4}{|c}{\bfseries GA-FPN}  \\
  \hline\hline
  P4 &   \checkmark & \checkmark & \checkmark & \checkmark  & \checkmark & \checkmark & \checkmark & \checkmark  \\
  P3 &    & \checkmark & \checkmark & \checkmark   &  & \checkmark & \checkmark & \checkmark \\
  P2 &    &  & \checkmark & \checkmark   &  &  & \checkmark & \checkmark  \\
  P1 &    &  &  & \checkmark   &  &  &  & \checkmark \\
  \hline
  CMP &   0.680 & \bfseries 0.696 & 0.654 & 0.473 & \bfseries 0.721 & 0.703 & 0.672 &0.554\\
  \end{tabular}}
\end{table}

\subsection{Comparison with State-of-the-art.}
Extensive experiments were performed on our large CT dataset. We mainly compared our approach with current state-of-the-art methods for object detection in computer vision fields such as RCNN \cite{ren2015faster, he2016deep, xie2017aggregated}, YOLO \cite{redmon2016you} and SSD \cite{liu2016ssd} as well as current state-of-the-art method for PNs detection. The results are mainly summarized in Tabele~3 and the other detail components can be found as follow. From Table~3 \footnote{In the abbreviation of the table: Calc. represents calcified nodules; Pleu. represents nodules on the pleura; 3-6, 6-10, 10-30 represents the longest diameter of solid nodules (mm); Mass. represents the case of solid nodules' longest diameter larger than 30 mm; p.ggn denotes pure GGN and m.ggn denotes mix ggn, or sub-solid nodules.}, we can observe that our system has achieved the highest CPM (0.733) with the fewest false positives rate (0.89) among this systems, which verifies the superiority of the improved GA-SSD in the task of lung nodule detection. On the classes of \emph{p.ggn} and \emph{m.ggn}, which are challenging to detect in clinical practice, our GA-SSD outperforms other approaches by a large margin.

To better justify the effectiveness of the proposed method, we conduct experiments over the LIDC-IDRI dataset \cite{armato2011lung} and obtained the competitive result with the state-of-the-art (Wang18 \cite{inbook}) method (CPM scores: 0.863 vs 0.878).

\begin{table}[htbp]
\begin{center}

\caption{Ablation study with the RCNN series, YOLO and SSD series on our chest CT dataset. The entries SSD300 used the input image resolution as the $300*300$ with the backbone of ResNeXt, and we use the SSD512 without bells and whistles as the baseline. FP rate represents the ratio of false positive (FP) to true positive (TP). Detailed information on the eight classes can be found in footnote 3.}

\setlength{\tabcolsep}{0.5mm}
\begin{tabular}{l | c | c | p{10mm} p{10mm} p{10mm} p{10mm} p{10mm} p{10mm}p{10mm}p{10mm}p{10mm}p{10mm}}
\hline
\bfseries Method  & \bfseries CPM & FP rate & Calc. & Pleu. & 3-6 & 6-10 &  10-30  & Mass & p.ggn & m.ggn\\
\hline\hline

RCNN \cite{ren2015faster}   &  0.464   & 1.30 & 83.8            & 55.6           & 77.4      & 90.5            &   84.4        & 77.8   & 83.9 & 89.7\\
RCNN  \cite{he2016deep}&  0.517   & 1.17 &  89.1            & 62.9           & 81.3      & 94.6            &   93.8        & 100   & 83.2 & 91.2\\
RCNN \cite{xie2017aggregated} &   0.538 & 0.99             & 86.9            & 62.4           & 78.9      & 91.9            &   93.8        & 100   & 86.1 & 92.6\\
\hline
SSD300  \cite{liu2016ssd} &     0.492   & 1.28&  91.0            & 68.4           & 84.7      & 90.5            &   93.8        & 100   & 86.9 & 92.6\\
SSD512  \cite{liu2016ssd} &     0.533   & 1.21 & 91.0            & 63.2           & 81.0      & 91.9            &   96.9        & 100   & 78.8 & 85.3\\
YOLO  \cite{redmon2016you} &     0.499   & 1.30 & 90.6            & 65.2           & 81.4      & 91.5            &   93.8        & 100   & 86.3 & 90.9\\
\hline
SSD300(ResNeXt)&   0.546   &  1.15 & 92.2            & 65.0           & 84.0      & 85.1            &   93.8        & 100   & 73.7 & 70.6\\
SSD512(ResNeXt)&   0.555   & 1.35 & 92.2            & 65.5           & 84.2      & 87.8            &   96.9        & 100   & 85.4 & 85.3\\
3DCNN\cite{inbook} &   0.700   &  1.59 & 91.3            & 60.3           & 80.5      & 91.9            &   93.8        & 100   & 85.0 & 91.2\\
GA-SSD512(ours) &  \bfseries 0.733 & \bfseries 0.89&  90.7            & 65.0           & 82.7      & 93.2            &   93.8        & 100   & \bfseries 94.2 & \bfseries 97.1\\
\hline
\end{tabular}
\end{center}
\end{table}

\section{Conclusion}

In this paper, we proposed a novel group-attention module based on 3D SSD with ResNeXt as the backbone for pulmonary nodule detection with CT scans. The proposed model showed superior sensitivity and fewer false positives compared to previous frameworks. Note that the higher sensitivity obtained, the more false positive pulmonary nodules resulted. Our architecture was shown to tackle the problems of high false positive rate caused by improving recall.

In the lung cancer screening step, radiologists will generally take a long time to read and analyze CT scans to make the correct clinical interpretation. But there are many factors making experienced radiologist prone to misdiagnosis, such as multi-sequence /multi-modality of images, the tiny size and low density of some lesions (such as GGN) that signal early lung cancer, heavy workload, and the repetitive nature of the job. Our proposed CNN-based system for pulmonary nodules detection achieve state-of-the-art performance with low false positives rate. Moreover, our proposed model takes only nearly 30s to detect pulmonary nodules, and it still has the potential to further speed up the detection process when more computing resources are available.


\bibliography{ma19}

\newpage
\appendix
\section{Sample results of the detection model}
\begin{figure}[!htb]
	\centering
	\includegraphics[width=12cm,height=11cm]{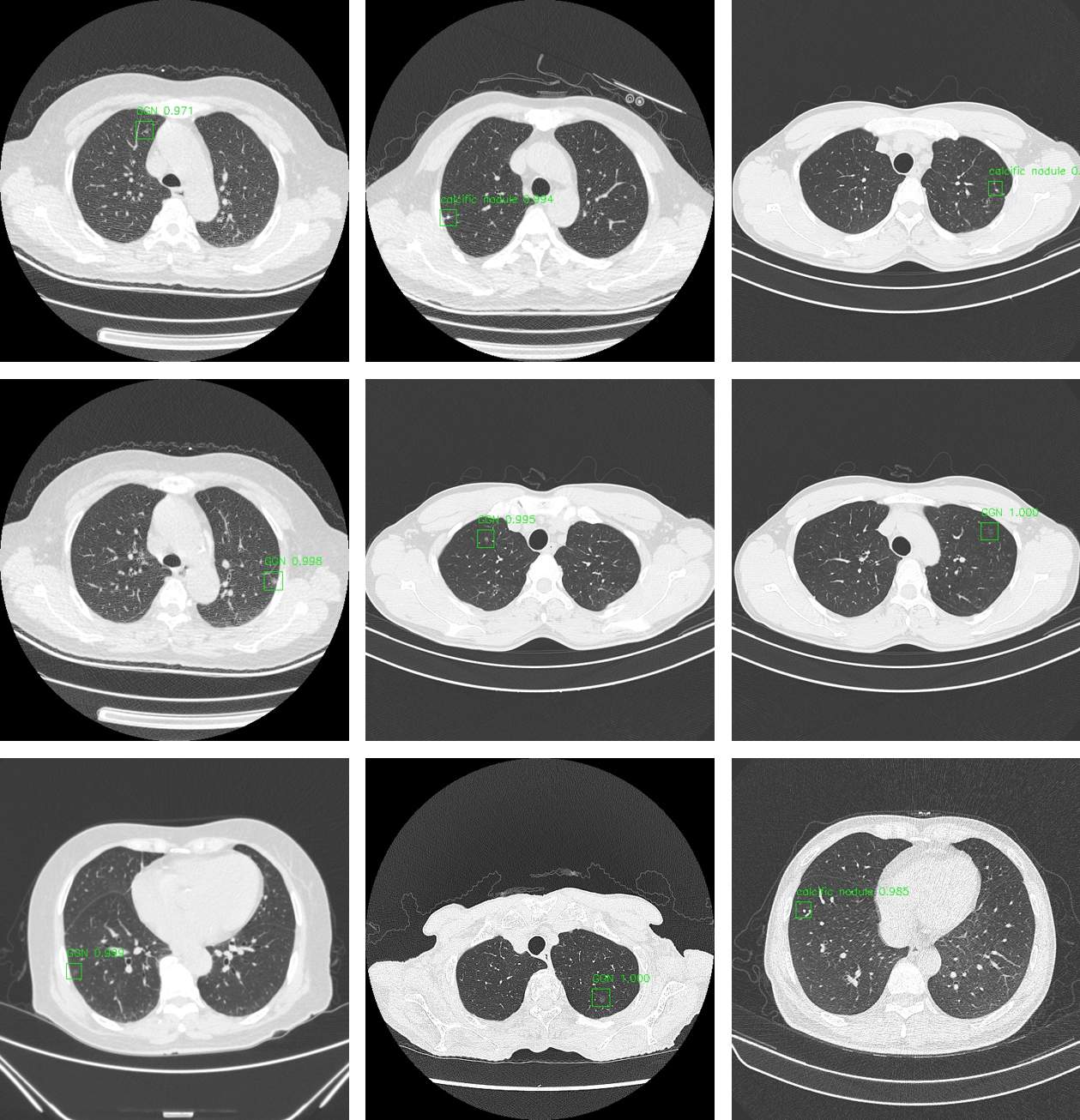}
	\caption{Results of true positives on nine cases. These nodules including the ones with small size are difficult to identify but are detected by our model. }
	\label{fig:nodule}
\end{figure}

\begin{figure}[!htb]
	\centering
	\includegraphics[width=12cm,height=4cm]{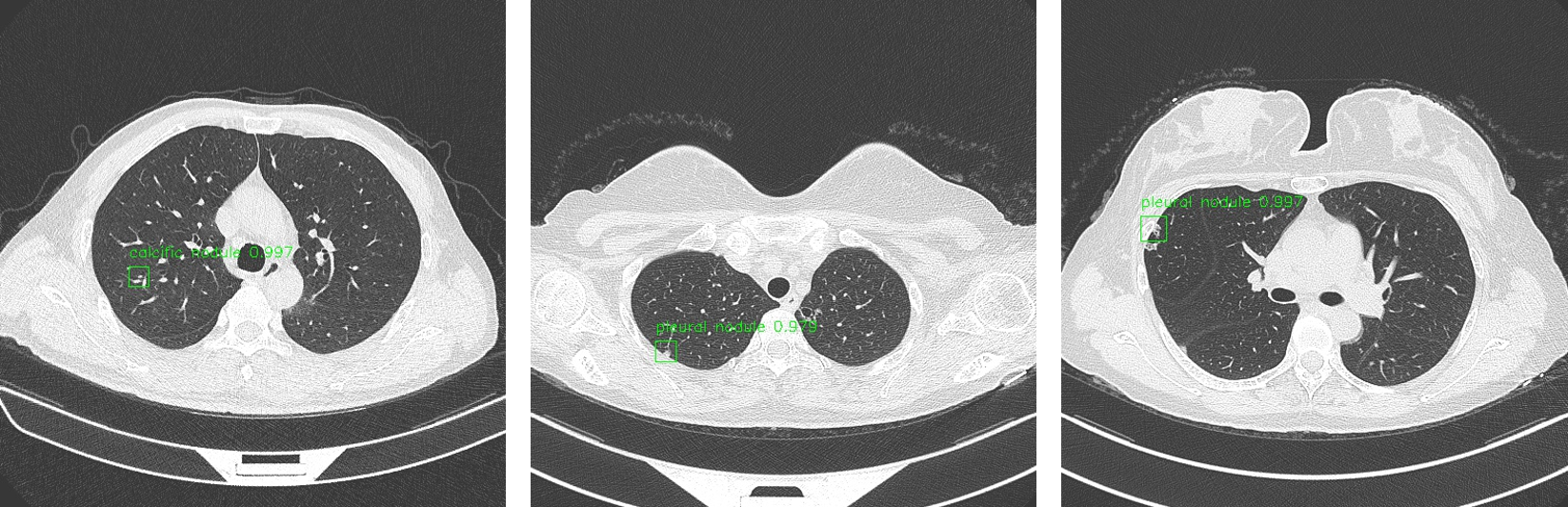}
	\caption{Results of false positives on three cases. These false positives have similar appearances with the nodules and are easily detected as abnormalities.}
	\label{fig:nodule1}
\end{figure}

\end{document}